\documentclass[a4paper]{article}
\usepackage{interspeech2013,amssymb,amsmath,graphicx,url}
\ninept

\title{Likelihood-ratio calibration using prior-weighted proper scoring rules}

\makeatletter
\def\name#1{\gdef\@name{#1\\}}
\makeatother
\name{{\em Niko Br\"{u}mmer$^1$ and George Doddington$^2$}}

\address{$^1$AGNITIO Research, Somerset West, South Africa \\
               $^2$Walnut Creek, CA, USA\\
{\small \tt niko.brummer@gmail.com, george.doddington@comcast.net}}

\def\model{\mathcal{M}}

\def\tar{{\tt{tar}}}
\def\non{{\tt{non}}}
\def\acc{\tt{accept}}
\def\rej{\tt{reject}}

\def\Cmiss{C_\text{miss}}
\def\Cfa{C_\text{fa}}

\DeclareMathOperator*{\argmin}{arg\,min}

\begin{document}
\maketitle

\begin{abstract}
Prior-weighted logistic regression has become a standard tool for calibration in speaker recognition. Logistic regression is the optimization of the expected value of the logarithmic scoring rule. We generalize this via a parametric family of proper scoring rules. Our theoretical analysis shows how different members of this family induce different relative weightings over a spectrum of applications of which the decision thresholds range from low to high. Special attention is given to the interaction between prior weighting and proper scoring rule parameters. Experiments on NIST SRE'12 suggest that for applications with low false-alarm rate requirements, scoring rules tailored to emphasize higher score thresholds may give better accuracy than logistic regression.
\end{abstract}

\section{Introduction}
We are concerned with the problem of speaker detection, where an automatic speaker recognizer is used to decide whether or not the voice of a designated target speaker is present in a given speech segment. Current speaker recognition algorithms output uncalibrated \emph{scores}, which have to be processed through a calibration stage before cost effective decisions can be made. This paper deals with the problem of designing the calibration stage. 

We implement the calibration stage in the form of a discriminative model, say $\model$, which outputs a posterior probability that the target speaker spoke the speech segment: $P(\tar|s,\model)$. The trial score, $s$, is computed by the speaker recognizer as a function of the target speaker and the speech segment. This posterior can be used in a straight-forward, standard way to make minimum-expected-cost Bayes decisions.

We shall take as the baseline discriminative model prior-weighted logistic regression~\cite{STBU}, which has become a standard recipe for calibration in speaker recognition, with implementations available in the FoCal~\cite{focal} and BOSARIS~\cite{BOSARIS} toolkits. 

Logistic regression is the minimization of the expected value of a special cost function, known as the \emph{logarithmic proper scoring rule}. We are interested here in generalizing this recipe by modifying the cost function. Our motivation derives from~\cite{CllrM10}, where it was demonstrated that a modified logistic regression that ignores scores below a suitable threshold can benefit applications with low false-alarm rate requirements. In this work we limit ourselves to cost functions which are \emph{proper scoring rules}~\cite{gneiting}. We expand on our previous work~\cite{talk}, to demonstrate theoretically and experimentally that we can tailor proper scoring rules to target the low false-alarm region.

\section{Proper scoring rules}
Given a database of supervised trials, the sum over trials of a proper scoring rule forms an objective function that can be used to simply evaluate the goodness of a recognizer with probabilistic output, and indeed also to facilitate discriminative training of such recognizers.

We restrict ourselves to \emph{binary} proper scoring rules, a family of special cost functions of the form $C^*(q,h)$, which evaluates the goodness of the recognizer output $q$, for a trial where hypothesis $h$ is true. We use the notation $h=\tar$ for a trial where the target speaker spoke and $h=\non$ for one where some other speaker spoke. In what follows it will be convenient to work with recognizer outputs in the form of posterior probabilities: $q=P(\tar|s,\model)$, where $s$ is the uncalibrated trial score and $\model$ is the calibration model. Later we show how to adapt this to recognizer outputs in likelihood-ratio form. 

\subsection{Definition}
\def\Dset{\mathcal{D}}
A binary proper scoring can be seen as a \emph{model of an application} of a detector---or more generally, a mixture of such applications. An application, $a$, is represented by a cost function $C_a(d,h)$, which maps outcomes to real-valued consequences. The outcome is composed of the decision, $d\in\Dset_a$, and the true hypothesis, $h\in\{\tar,\non\}$. The sets from which decisions are chosen may differ between applications, for example $\Dset_a=\{\acc,\rej\}$, or $\Dset_a=\{\acc,\rej,{\tt undecided}\}$, and so on. Our application model assumes that the detector output, $q$, is used to make a minimum-expected-cost Bayes decision:\footnote{We assume there is a rule---the details of which are unimportant here---to choose among multiple minimizers.} 
\begin{align*}
d^*_a(q)\in\argmin_d qC_a(d,\tar)+(1-q)C_a(d,\non)
\end{align*} 
The associated \emph{proper scoring rule} is defined as the cost of this decision~\cite{degroot}:
\begin{align}
\label{eq:defpsr}
C^*_a(q,h)&=C_a(d^*_a(q),h)
\end{align}
A convex combination of proper scoring rules is still a proper scoring rule, in the sense that it can be derived via~\eqref{eq:defpsr} from a suitably constructed cost function that represents a mixture of different applications~\cite{msli}.

\subsection{Canonical form}
Although applications may be defined via a large variety of cost functions, all of this variety can be conveniently represented in a surprisingly simple form~\cite{msli,gneiting,degreesofboosting}. All binary proper scoring rules can be expressed in the form:\footnote{Regularity conditions apply to the cost functions. We have adapted the representation in~\cite{msli}, section 7.4.1, via a transformation to log odds domain.}
\begin{align}
\label{eq:canonical}
\begin{split}
C^*_w(q,\tar) &= k_0\int_{\log\frac{q}{1-q}}^\infty (1+e^{-t})w(t)\;dt +k_1 \\
C^*_w(q,\non) &= k_0\int_{-\infty}^{\log\frac{q}{1-q}} (1+e^t)w(t)\;dt +k_2
\end{split}
\end{align}
where $k_0>0$, $w(t)\ge0$ and $\int_0^1w(t)\;dt=1$. The constants $k_0,k_1,k_2$ don't play any useful role in the Bayes decision framework and may be set to $k_0=1$ and $k_1=k_2=0$ without loss of generality~\cite{degroot}. 

The weighting distribution $w(t)$ can be very general, including smooth functions, step functions or even impulses. The impulse $w(t)=\delta(t-\theta)$ represents a single, \emph{simple application}, with binary decisions in $\Dset_a=\{\acc,\rej\}$, a cost function with penalties %$\Cmiss=C_a(\rej,\tar)=\frac{1}{\theta}$ and $\Cfa=C_a(\acc,\non)=\frac{1}{1-\theta}$ 
$\Cmiss=1+e^{-\theta}$ and $\Cfa=1+e^\theta$ 
and the Bayes decision threshold at $\log\frac{q}{1-q}=\theta$. A sum of impulses represents a discrete mixture of applications, while a smooth function represents a continuous mixture over a continuous spectrum of applications. The important point here is that all Bayes decision applications (or mixtures of applications) can be represented by~\eqref{eq:canonical} via a suitable choice of $w(t)$.

\section{Objective function}
\def\Cbar{\bar{C}}
\def\Tset{\mathcal{T}}
\def\Nset{\mathcal{N}}
Here we build an objective function, for training or evaluation, out of a proper scoring rule, $C^*_w$. The proper scoring rule models the cost of making a Bayes decision in a single detection trial. To turn this into an objective function, we take the \emph{expected} cost over a whole  supervised database. We form this expectation via two hypothesis conditional averages, weighted by a synthetic class prior (usually different from the database proportions), of the form $\pi=P(\tar)$, $1-\pi=P(\non)$. The expected cost is:
\begin{align}
\label{eq:obj1}
\Cbar_w^\pi &= \frac{\pi}{T}\sum_{i\in\Tset} C^*_w(q_i,\tar)
+\frac{1-\pi}{N}\sum_{i\in\Nset} C^*_w(q_i,\non)
\end{align}
where $q_i$ is the recognizer's posterior for trial $i$; $\Tset$ is a set of $T$ target trial indices; and $\Nset$ a set of $N$ non-target indices.

\subsection{Scoring likelihood-ratios}
Now, we relieve the calibrator of the (implicit) responsibility to have a hypothesis prior and to produce posteriors. Instead, we require it to output \emph{log-likelihood-ratios}, denoted $\ell_i$. Since the prior is already fixed as a parameter of the objective~\eqref{eq:obj1}, the recognizer's posterior is given by Bayes' rule as:\footnote{In our discriminative calibration framework, defining $\ell_i$ as the ratio of two generative likelihoods adds no value. Instead~\eqref{eq:bayes} should be taken as the definition of $\ell_i$.}
\begin{align}
\label{eq:bayes}
q_i = \sigma(\ell_i+\tau)
\end{align}
where $\tau$ is prior log odds and $\sigma$ is the logistic sigmoid:
\begin{align}
\tau &= \log\frac{\pi}{1-\pi}\;,&\text{and}&&
\sigma(x)&=\frac{1}{1+e^{-x}}
\end{align}
After some manipulation, we can express the functions of $q_i$ in~\eqref{eq:obj1} as functions of $\ell_i$ instead: 
\begin{align}
\label{eq:manip}
\begin{split}
\pi C^*_w(q_i,\tar) &= \int_{\ell_i}^\infty (1+e^{-t})w_\pi(t)\;dt \\
(1-\pi)C^*_w(q_i,\non) &= \int_{-\infty}^{\ell_i} (1+e^t)w_\pi(t)\;dt
\end{split}
\end{align}
where $w_\pi(t)=r_\tau(t)w(t+\tau)$ is a translated and modulated version of $w(t)$, where the modulation factor:
\begin{align}
\label{eq:rtau}
r_\tau(t) &= \frac{1+e^{t+\tau}}{(1+e^t)(1+e^\tau)}
\end{align}
is a raised and scaled sigmoid. If $w(t)$ is normalized, then $w_\pi(t)$ is normalizable in the sense: $0<\int_{-\infty}^{\infty}w_\pi(t)\;dt<1$.

\subsection{Averaging over $\pi$}
The integrals in~\eqref{eq:canonical} can be interpreted as averaging over all possible cost functions, where $w(t)$ determines the relative importance of different cost functions. In our construction of the objective function~\eqref{eq:obj1}, we inherit this averaging over cost functions, but we seem to have fixed the prior at $\pi$. Since different applications of a speaker detector can be expected to have different priors, should we not also average over $\pi$? We can, but it adds no generality, because this would effectively just replace $w_\pi(t)$ in~\eqref{eq:manip} by some other normalized/normalizable weighting distribution, say $\Omega(t)$. We can now reformulate our objective function in its most general form as:
\begin{align}
\label{eq:obj2}
\begin{split}
\Cbar_\Omega &= \frac{1}{T}\sum_{i\in\Tset} C_\Omega(\ell_i,\tar)
+\frac{1}{N}\sum_{i\in\Nset} C_\Omega(\ell_i,\non) 
\end{split}
\end{align}
where
\begin{align}
\label{eq:omegaint}
\begin{split}
C_\Omega(\ell_i,\tar) &= \int_{\ell_i}^\infty (1+e^{-t})\Omega(t)\;dt \\
C_\Omega(\ell_i,\non) &= \int_{-\infty}^{\ell_i} (1+e^t)\Omega(t)\;dt
\end{split}
\end{align}
We now have a general objective function, parametrized by $\Omega(t)$, which effectively determines the relative weighting over a mixture of different applications, each of which can have a different cost function and target prior. Notice that $t$ is just the log-likelihood-ratio threshold. Simple $\acc$/$\rej$ applications use a single threshold, while more complex applications (or mixtures) have multiple thresholds. Thus, the combination of application costs in~\eqref{eq:omegaint} is accomplished via representing applications by their thresholds.

\section{Practical recipe}
Here we introduce a practical choice for the weighting distribution, $\Omega(t)$ and show how it is used for calibration. Since calibration involves optimization, smooth differentiable distributions are easier to work with. It is also desirable that the integrals have closed-form solutions. To this end, we adopt a 2-parameter (here $\alpha,\beta$) family of proper scoring rules proposed in~\cite{degreesofboosting}. This family is effectively augmented by a third parameter, $\tau=\log\frac{\pi}{1-\pi}$, via the prior-weighting in~\eqref{eq:obj1}. This gives $\Omega=\Omega_{\alpha,\beta,\tau}$:
\begin{align}
\label{eq:omegadef}
\Omega_{\alpha,\beta,\tau}(t) &= \frac{r_\tau(t) w_{\alpha,\beta}(t+\tau)}{Z_{\alpha,\beta,\tau}}
\end{align}
where $\alpha,\beta>0$; $r_\tau$ is defined in~\eqref{eq:rtau}; $Z_{\alpha,\beta,\tau}$ ensures the distribution is normalized; and $w_{\alpha,\beta}$ is the beta distribution, transformed to log-odds domain:
\begin{align}
w_{\alpha,\beta}(t) &= \frac{\sigma(t)^\alpha\sigma(-t)^\beta}{B(\alpha,\beta)}
\end{align}
and $B(\alpha,\beta)$ is the beta function. When $\alpha,\beta\in\{\frac12,1,1\frac12,2,...\}$, then using $w=w_{\alpha,\beta}$ in~\eqref{eq:canonical} gives closed-form solutions, which can be plugged into~\eqref{eq:obj1}, with $\pi=\sigma(\tau)$ and $q_i=\sigma(\ell_i+\tau)$. 

\subsection{Interpretation}
\begin{figure}[t]
\centerline{\includegraphics[width=0.5\textwidth,trim = 2cm 1.8cm 2cm 2cm,clip = true]{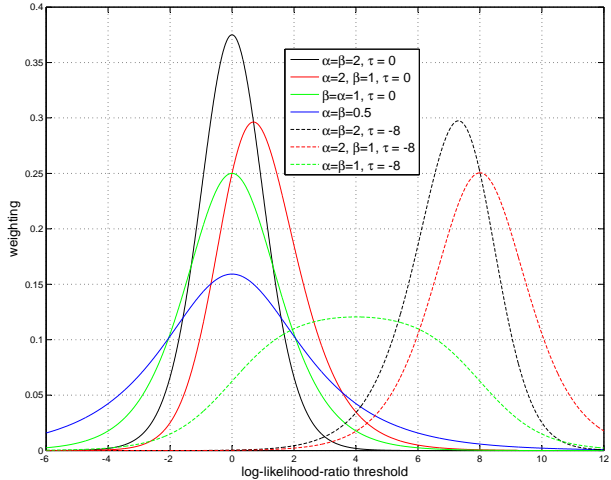}}
\caption{Threshold weightings, $\Omega_{\alpha,\beta,\tau}(t)$, of a few objective function parametrizations. (Solid green represents $C_\text{llr}$, see main text.)}  
\label{fig:omega}
\end{figure}

While this recipe is practically realized via~\eqref{eq:canonical} and~\eqref{eq:obj1}, a theoretical interpretation is given by~\eqref{eq:omegadef} and~\eqref{eq:obj2}, because for this parametrization we have $\Cbar_w^\pi=Z_{\alpha,\beta,\tau}\Cbar_\Omega$. We exploit this in figure~\ref{fig:omega}, where we plot~\eqref{eq:omegadef} for several values of the parameters. Parameters $\alpha,\beta$ control left and right tails respectively---when they are increased tails become thinner and the distributions become narrower (higher modes). Location is indirectly controlled by $\tau$, but this depends on $\alpha,\beta$. For the case $\alpha=\beta=\frac12$, the distribution is \emph{invariant} to $\tau$. For $\alpha=\beta=1$, the mode shifts only as far as $-\frac{\tau}{2}$, while the distribution is considerably widened. However, when $\alpha=2$, the mode shifts close to $-\tau$ and the widening is less dramatic. We show in our experiments below that when we target applications with high thresholds (low false-alarm rate), then the effective shifting allowed by $\alpha=2$ leads to better performance than the halfway shifted and stretched version given by the baseline logistic regression solution with $\alpha=\beta=1$.

\subsection{Examples}
Here we list some solutions for the integrals in~\eqref{eq:canonical}, which we used in our experiments. The \emph{boosting rule}~\cite{degreesofboosting}, with $\alpha=\beta=\frac12$:
\begin{align}
\begin{split}
C^*_{\frac12,\frac12}(q,\tar)&=\frac{2}{\pi}\sqrt\frac{1-q}{q} \\
C^*_{\frac12,\frac12}(q,\non)&=\frac{2}{\pi}\sqrt\frac{q}{1-q} 
\end{split}
\end{align}
The \emph{logarithmic rule}, with $\alpha=\beta=1$, forms the objective function for logistic regression and is our baseline:
\begin{align}
\begin{split}
C^*_{1,1}(q,\tar)&=-\log(q) \\
C^*_{1,1}(q,\non)&=-\log(1-q) 
\end{split}
\end{align}
The parametrization $\alpha=\beta=1, \tau=0$ gives (up to scaling), the objective $C_\text{llr}$, proposed in~\cite{focal} for the evaluation of goodness of recognizers with likelihood-ratio output. The \emph{Brier rule}~\cite{brier}, with $\alpha=\beta=2$:
\begin{align}
\begin{split}
C^*_{2,2}(q,\tar)&=3(1-q)^2 \\
C^*_{2,2}(q,\non)&=3q^2 
\end{split}
\end{align}
An \emph{asymmetric} rule, with $\alpha=2,\beta=1$:
\begin{align}
\begin{split}
C^*_{2,1}(q,\tar)&=2(1-q) \\
C^*_{2,1}(q,\non)&=-2\log(1-q)-2q 
\end{split}
\end{align}

\subsection{Calibration recipe}
We assume that our recognizers output uncalibrated scores. Let the score for a trial $i$ be denoted $s_i$. We then apply a parametric (affine) calibration transformation:
\begin{align}
\label{eq:caltrans}
\ell_i &= A s_i + B
\end{align}
The calibration parameters $A,B$ need to be trained on a calibration training database. For training purposes, we choose and fix the parameters $\alpha,\beta,\tau$ and then minimize the objective~\eqref{eq:obj1} w.r.t.\ $A,B$. In the special case $\alpha=\beta=1$ such training is known as logistic regression. Since the objective and the calibration transformation are differentiable, one can obtain the gradient w.r.t.\ $A,B$ by backpropagation and then use any of a variety of well-known unconstrained numerical optimization algorithms. For this work, we used  BFGS~\cite{nocedal}.

\section{Experiments}
\def\Cprim{C_\text{primary}}
We performed calibration experiments on scores from a single speaker recognizer (an i-vector PLDA system), which was part of the ABC submission~\cite{abc} to the NIST SRE'12 speaker recognition evaluation. Several calibration transformations of the form~\eqref{eq:caltrans} were trained (separately for males and females) on a large development set with multiple microphone and telephone speech segments of 2019 speakers from SRE'04, '05, '06, '08 and '10. This gave about 120 million scores for calibration training. The different calibration transformations were obtained by using different objective function parameters, $\alpha,\beta,\tau$. 

We tested the goodness of these calibrators on the NIST SRE'12 extended trial set~\cite{SRE12evalplan}, where we pooled males and females and all 5 common evaluation conditions, giving about 80 million trials. 

Our evaluation criterion was $\Cprim$, with $P_\text{known}=0$, as defined in~\cite{SRE12evalplan}. It should be noted that this criterion can \emph{also} be interpreted as an example of our objective function~\eqref{eq:obj2}, since $\Cprim=\Cbar_\Omega$, if we choose $\Omega=\Omega_\text{primary}$, defined as:
\begin{align}
\Omega_\text{primary}(t) &= \frac12\sum_{i=1}^2 \sigma(\theta_i) \delta(t-\theta_i)
\end{align}
which is concentrated in two impulses\footnote{with \emph{almost} equal weights, since $\sigma(\theta_i)\approx1$} at the log-likelihood-ratio threshold values of $\theta_1=4.59$ and $\theta_2=6.91$. This criterion was chosen by NIST to be calibration-sensitive, but only in the `low false-alarm region' around these operating points. 

\begin{figure}[t]
\centerline{\includegraphics[width=0.5\textwidth,trim = 2cm 1.8cm 2cm 2cm,clip = true]{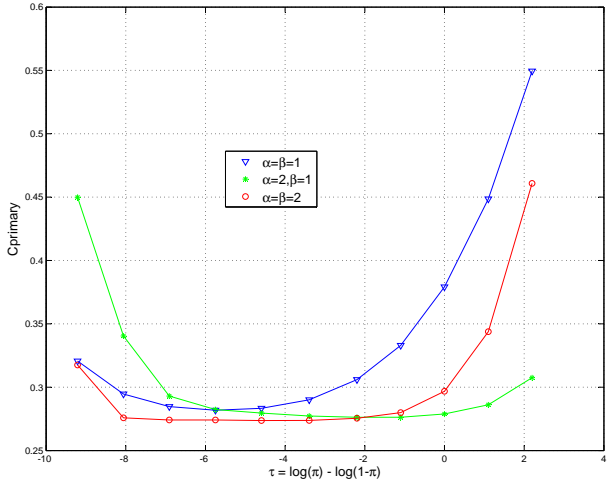}}
\caption{SRE'12: Comparison between proper scoring rules for calibration. More negative $\tau$ places more weight at more positive score thresholds. (Blue at $\tau=0$ represents $C_\text{llr}$.)}  
\label{fig:results}
\end{figure}

Our experimental results are summarized in figure~\ref{fig:results}, where we plotted $\Cprim$ obtained by using the calibrators formed by three representative proper scoring rules against the prior-weighting parameter, $\tau=\log\frac{\pi}{1-\pi}$. Our baseline (blue triangles) is logistic regression ($\alpha=\beta=1$). The other two rules with $\alpha=2$ (red circles, green asterisks), gave better performance in the sense of having lower, wider minima.

The boosting rule ($\alpha=\beta=\frac12$) performed very poorly, at $C_\text{primary}=0.65$, and is not shown on the graph. Its thick tails makes it vulnerable to outliers.

\section{Discussion}
The reason why different proper scoring rules perform differently is that the affine calibration transformation is limiting. It cannot satisfy the requirement of good calibration at all operating points simultaneously. We have to choose, via $\Omega(t)$, which operating points are important and which can be ignored. This is illustrated in figure~\ref{fig:pav}, where we compare the log-likelihood-ratios produced for SRE'12 by two of the proper scoring rule calibrations (horizontal axis), against ideal, reference log-likelihood-ratios (vertical axis). Ideally these plots should lie along the identity transform (the black diagonal through the origin). But the affine calibration transformation can only try to fit these curves to the ideal straight line via scaling and shifting the curves along the horizontal axis. The logistic regression ($\alpha=\beta=1$) places more weight at lower threshold values and indeed manages to be closer to the ideal there, to the detriment of calibration at the higher $C_\text{primary}$ operating points. On the other hand, the Brier rule ($\alpha=\beta=2$), which places more weight at the $C_\text{primary}$ operating points, does better there, but pays for this by doing worse at lower threshold values. 

\subsection{The PAV reference}
\begin{figure}[!t]
\centerline{\includegraphics[width=0.5\textwidth,trim = 4cm 1.8cm 4cm 2cm,clip = true]{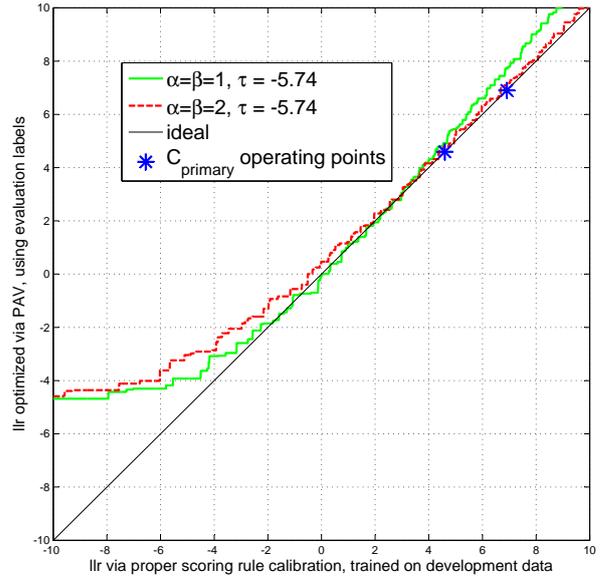}}
\caption{SRE'12: PAV reference vs proper scoring rule calibration}  
\label{fig:pav}
\end{figure}

The ideal reference along the vertical axis of figure~\ref{fig:pav} is achieved via an optimization algorithm known as \emph{pool adjacent violators} (PAV)\cite{pav,msli}. This algorithm finds the optimal log-likelihood-ratio value for every trial non-parametrically, subject only to the constraint that when sorted along the real line, the order of the optimized values must be the same as the sorted order of the original uncalibrated scores. This is equivalent to constraining the calibration transformation to be monotonically rising. The PAV output is optimal in the sense that it simultaneously optimizes all binary proper scoring rules at all values of the prior weighting~\cite{msli}. The PAV solution can be optimal everywhere because the non-parametric monotonicity constraint is less strict than the parametric affine transformation available to the proper scoring rule calibrators. It is important to note however that the true SRE'12 class labels were used to perform the PAV optimization, while the proper scoring rule calibrations were optimized on the independent development data. If the PAV had been optimized on the development data, it would no longer have this optimality on the evaluation data.\footnote{We \emph{did} interpolate the PAV transform of the development scores to form a calibrator for the evaluation data. This gave $C_\text{primary}$ very similar to the best proper scoring rule results shown in figure~\ref{fig:results}. The details are out of scope for this paper.}

\subsection{Caveat}
Conclusions from our experimental results should not be extrapolated without caution. The superiority demonstrated here of rules with $\alpha>1$ may not hold for: (i) small calibration training databases; (ii) other recognizers; (iii) applications with thresholds in other operating regions. 

\subsection{Generalization}
The objective function family presented here could be more generally applied, not just for calibration. These objectives could be used for fusion of multiple recognizers, or indeed for more general discriminative training of speaker recognizers in the manner of~\cite{dplda}. However for more complex recognizers, the risk of overtraining is greater---and this risk may be compounded by more narrowly focussed objectives, such as the Brier rule. In contrast, the wider focus of the logarithmic rule has a regularizing effect, which combats overtraining.

\clearpage
\eightpt
\bibliographystyle{IEEEtran}

\end{document}